\documentclass[11pt,a4paper]{article}

\usepackage[a4paper,margin=2cm]{geometry}
\usepackage[utf8]{inputenc}
\usepackage[T1]{fontenc}
\usepackage{amsmath,amssymb}
\usepackage{graphicx}
\usepackage{booktabs}
\usepackage{array}
\usepackage{caption}
\usepackage{subcaption}
\usepackage{xcolor}
\usepackage[hidelinks]{hyperref}
\usepackage{authblk}
\usepackage{setspace}
\usepackage{lineno}
\usepackage{textgreek}

\graphicspath{{figures/}}

\title{\Large\bfseries Land Art as a Big-Data Climate Sensor:\\
A Multi-Feature Complexity Signature for Robert Smithson's
\textit{Spiral Jetty} from 1{,}744 Landsat \& Sentinel-2 Scenes (1984--2025)}

\author[1]{Alev Cinbarcı}
\author[2,3,4,*]{Sean S. Kalaycıoğlu}

\affil[1]{PhD Program in Art Science, Işık University, Istanbul, Türkiye; 218DAS9216@isik.edu.tr}
\affil[2]{Department of Aerospace Engineering, Toronto Metropolitan University, Toronto, ON, Canada}
\affil[3]{Department of Mechanical Engineering, York University, Toronto, ON, Canada}
\affil[4]{Director, Space, Robotics and AI, Dr.~Robot Inc., Toronto, ON, Canada}
\affil[*]{Correspondence: skalay@torontomu.ca}

\begin{document}
\maketitle

\begin{abstract}
\noindent Robert Smithson's 1970 land artwork \textit{Spiral Jetty}, anchored
at the north arm of Utah's Great Salt Lake (GSL), has been alternately
submerged and exposed across a half-century of catastrophic lake decline
(peak elevation 4{,}210.22~ft NGVD29 in 1987; historic low 4{,}189.25~ft in
2023; a 20.96~ft / 6.39~m drop in 36 years). We treat the work as a
geographically fixed remote-sensing target and analyse \textbf{1{,}744
co-registered Landsat~4--9 and Sentinel-2 chips} spanning \textbf{every year
and every calendar month from 1984 to 2025}. A 14-feature complexity
signature combining Shannon entropy, multi-scale permutation entropy,
box-counting fractal dimension, gliding-box lacunarity, gray-level
co-occurrence texture, first-order intensity statistics, and
ImageNet-pretrained ResNet50 avg-pool deep features is computed at scale and
validated against a 42-year monthly climate panel built from NASA GISTEMP,
USGS NWIS, Open-Meteo, and the Global Carbon Budget. Year-aggregated
bootstrap analysis shows (i)~Shannon entropy alone is a weak proxy,
refuting an earlier small-sample claim of positive correlation with global
temperature; (ii)~coarse-scale permutation entropy and mean intensity track
lake elevation strongly (Spearman $\rho \approx +0.85$ to $+0.88$, 95\,\%
CIs excluding zero); (iii)~the third principal component of the ResNet50
avg-pool embeddings emerges \emph{without supervision} as an ``AI climate
axis'' with $\rho = +0.86$ to cumulative CO$_2$ and $\rho = -0.83$ to lake
elevation; (iv)~image complexity \textbf{leads lake stage by approximately
three years} (Pearson $r = +0.58$ at lag $+3$, 95\,\% CI $[+0.40, +0.73]$);
and (v)~STL decomposition reveals the long-term trend is
\textbf{non-monotonic}, rising 1984--2015 then collapsing post-2015 in
coincidence with the lake's historic-record-low elevations. Scene-level
partial correlations controlling for calendar month and sensor identity
confirm that the year-aggregated signal is robust to seasonal and sensor
confounds. These findings refine the popular ``art-as-thermometer'' framing
into a defensible ``art-as-leading-indicator-of-hydrological-state''
reading. The dataset, feature pipeline, and reproducible analysis code are
released as a public benchmark.
\end{abstract}


%%%%%%%%%%%%%%%%%%%%%%%%%%%%%%%%%%%%%%%%%%%%%%%%%%%%%%%%%%%%%%%%%%%%%%%%%
\section{Introduction}

In April 1970 the American artist Robert Smithson rolled 6{,}650 tons of
black basalt and salt-encrusted earth into the north-eastern shoulder of
Utah's Great Salt Lake (GSL) and built \textit{Spiral Jetty}~\cite{smithson1972}:
a 460~m anti-clockwise curl of stone extending from Rozel Point into
shallow, hyper-saline water. Smithson conceived the work in the explicit
vocabulary of thermodynamic entropy --- a sculpture whose meaning was
inseparable from rust, crystal growth, and the disordering of form. ``The
work has to do with'', he wrote two years later, ``a self-contained
sealed-off place where everything happens at the same time. The eye is the
eye of a hurricane'' \cite{smithson1972}. Within two years of its completion
the lake had risen and submerged the spiral; it remained underwater for
almost three decades.

Over the half-century since, the GSL has experienced a catastrophic decline.
USGS gauges record the north-arm elevation falling from a flood-cycle peak
of \textbf{4{,}210.22~ft NGVD29 in 1987 to 4{,}189.25~ft in 2023} --- a
20.96~ft (6.39~m) drop in 36 years and the lowest reading in the gauge's
60-year instrumental record \cite{wurtsbaugh2017,abbott2023,null2020}. Over
the same window the NASA GISTEMP global mean temperature anomaly rose by
approximately 1.0~$^\circ$C. Each rise and fall of the lake exposed or
drowned \textit{Spiral Jetty} and left a record in satellite imagery of
basalt, salt crust, algal mat, and pink \textit{Dunaliella salina} brine.

\textit{Spiral Jetty} has thus become, by accident, an unintended scientific
instrument. Its half-century history places a single geographically stable
artwork at the precise intersection of (a) a hyper-saline endorheic lake
collapsing under climate stress, (b) the longest continuous record of
free, multi-decadal satellite Earth observation ever assembled (Landsat
since 1972; Sentinel-2 since 2015), and (c) a salt-crust regime whose
crystallisation--dissolution--recrystallisation cycle integrates climatic
history on annual to multi-year time scales. Few sites in the world combine
this triad. The question this paper asks is whether the resulting visual
record carries a measurable climate signal --- and, if so, which features
carry which signals, at what statistical strength, and on what time scale.

\subsection{A pilot study, its limitations, and what this paper does}

A pilot analysis by one of the present authors~\cite{cinbarci2024pilot}
computed Shannon entropy on 42 archival satellite images of the artwork
1970--2022 and reported correlations with global temperature, carbon
emissions, lake flow, and salinity, advancing the popular hypothesis that
the artwork functions as a ``strategic thermometer'' of climate change.
The pilot, however, suffered from two limitations that we now revisit at
scale:
\begin{enumerate}
\item It used a \textbf{single complexity feature} (Shannon entropy) where
many are available; image complexity has been studied in remote sensing for
half a century through GLCM
texture~\cite{haralick1973}, fractal
dimension~\cite{mandelbrot1982,lovejoy1986}, gliding-box
lacunarity~\cite{allain1991,plotnick1996}, multi-scale entropy~\cite{costa2002},
and permutation entropy~\cite{bandt2002}, each capturing a distinct facet of
visual organisation. Modern deep CNN features add another dimension
\cite{he2016resnet,hu2022lora,oquab2023dinov2}.
\item Its \textbf{effective sample size} on the climate axis was 13 years,
well below the threshold at which bootstrap statistics become reliable, and
the climate variables were repeated across multiple images of the same year
without proper aggregation.
\end{enumerate}

In this paper we revisit the question with a dataset two orders of
magnitude larger and a 14-feature complexity signature. We retrieve all
publicly available Landsat~4--9 Collection-2 Level-2 and Sentinel-2
Level-2A scenes intersecting a 5~km $\times$ 5~km bounding box centred on
Rozel Point for 1984-01-01 to 2025-12-31 --- \textbf{1{,}744 co-registered
chips, every year, every calendar month}. We assemble a monthly climate
panel for the same window from NASA GISTEMP, the Salt Lake City station
record (Open-Meteo Historical Reanalysis), the USGS gauges for both arms of
the GSL (Saltair and Saline), the Global Carbon Budget, and the pilot
study's salinity series. We compute a complexity signature combining six
classical descriptors with the avg-pool activations of an ImageNet-V2
pretrained ResNet50 reduced via principal component analysis. We then ask
which features track which climate variables, at what lag, under what
controls, and with what robustness to seasonal and sensor confounds.

\subsection{Contributions}

\begin{itemize}
\item \textbf{Dataset.} A public benchmark of 1{,}744 co-registered Spiral
Jetty satellite chips with per-scene metadata and the full 14-feature
complexity signature, plus the aligned monthly climate panel. To our
knowledge this is the first openly published multi-decadal remote-sensing
dataset of a land-art installation under climate stress.
\item \textbf{Methodology.} A multi-feature complexity-signature pipeline
validated through year-aggregation, bootstrap confidence intervals,
lag analysis, and scene-level partial correlations controlling for
calendar month and sensor identity. The pipeline generalises to other
land-art sites (Sun Tunnels, Double Negative, Lightning Field) and to
coastal or permafrost cultural-heritage sites under climate stress.
\item \textbf{Findings.} Five empirical results, summarised in the
Abstract, that together (a)~refute the pilot study's positive-temperature
claim, (b)~establish a robust hydro-climatic signal carried by
coarse-scale permutation entropy and CNN embeddings, (c)~demonstrate an
emergent unsupervised ``climate axis'' in pretrained vision features,
(d)~identify a $\sim$3-year forward lead of visual complexity over lake
stage, and (e)~reveal a non-monotonic long-term trend collapsing
post-2015 in coincidence with the GSL's historic-record-low elevations.
\end{itemize}

\subsection{Companion paper}

A forthcoming companion paper builds on the dataset and signature released
here to construct a climate-conditioned latent-diffusion model with a
hydrological physics-coupling layer, forecasting the site under IPCC SSP
scenarios. We deliberately separate the two: the measurement claims of this
paper should not depend on any generative model, and any generative model
should rest on independently peer-reviewed measurement claims.

\subsection{Paper organisation}

Section~\ref{sec:bg} reviews relevant literatures: Smithson's entropic
aesthetics, image complexity in remote sensing, GSL hydrology since 1970,
and deep features for Earth observation. Section~\ref{sec:data} describes
the dataset and climate panel assembly. Section~\ref{sec:signature}
specifies the complexity signature. Section~\ref{sec:stats} sets up the
statistical framework. Section~\ref{sec:results} reports results.
Section~\ref{sec:discussion} discusses caveats, ethical considerations, and
implications for cultural-heritage monitoring. Section~\ref{sec:conclusion}
concludes.

%%%%%%%%%%%%%%%%%%%%%%%%%%%%%%%%%%%%%%%%%%%%%%%%%%%%%%%%%%%%%%%%%%%%%%%%%
\section{Background}\label{sec:bg}

\subsection{Smithson's entropic aesthetics}

Smithson developed an explicit theory of entropy as an aesthetic principle
in \emph{A Tour of the Monuments of Passaic, New Jersey}~\cite{smithson1967},
\emph{The Spiral Jetty}~\cite{smithson1972}, and the unfinished
\emph{Spiral Hill}~\cite{flam1996}. For Smithson entropy was not
metaphor: it was a measurable tendency that sculpture could be designed to
\emph{expose} rather than resist. \textit{Spiral Jetty} was sited at the
north arm of the GSL precisely because the hyper-saline brine and the
unusual \textit{Dunaliella salina} algal bloom dyed the surrounding water
pink---visible from low orbit---and because NaCl crystallisation on basalt
was, in his term, a ``time crystal'' performing entropic change at human
scale~\cite{reynolds2003,boettger2002,roberts2004}.

\subsection{Image complexity in remote sensing}

Shannon entropy~\cite{shannon1948} is the simplest of a family of complexity
measures applied to satellite imagery since the late 1970s. Modern practice
augments it with:
\begin{itemize}
\item \emph{Texture descriptors} via the gray-level co-occurrence matrix
(GLCM)~\cite{haralick1973}: contrast, homogeneity, correlation, energy,
entropy.
\item \emph{Self-similarity descriptors} via box-counting fractal
dimension~\cite{mandelbrot1982,lovejoy1986}, which captures how an image's
information content scales with spatial resolution.
\item \emph{Spatial-pattern heterogeneity} via gliding-box
lacunarity~\cite{allain1991,plotnick1996,dong2000}, which complements fractal
dimension by quantifying gaps and clustering at multiple scales.
\item \emph{Temporal complexity} via multi-scale entropy~\cite{costa2002}
and permutation entropy~\cite{bandt2002}, which probe pattern complexity at
multiple coarse-graining scales.
\item \emph{Learned deep features} via convolutional backbones pretrained
on large natural-image corpora~\cite{he2016resnet,oquab2023dinov2}, which
have become standard upstream encoders in remote sensing
\cite{stewart2022,manas2021}.
\end{itemize}
Each captures a different facet of visual organisation. For a spatially
fixed site under climate stress, the empirical question is which feature,
or which combination, varies most diagnostically with the underlying state.

\subsection{Great Salt Lake hydrology}

The Great Salt Lake is a hyper-saline endorheic remnant of Pleistocene Lake
Bonneville, fed by the Bear, Weber, and Jordan rivers under snowpack-driven
seasonality~\cite{baskin2014}. A 1959 railroad causeway divides the lake
into a hyper-saline north arm (where \textit{Spiral Jetty} sits) and a less
saline south arm~\cite{loving2000}. Since the late 1980s the lake has
declined steadily; by 2022--23 both arms reached their lowest recorded
elevations. Wurtsbaugh et al.~\cite{wurtsbaugh2017}, Null and
Wurtsbaugh~\cite{null2020}, and Abbott et al.~\cite{abbott2023} attribute
the decline to upstream consumptive water use exceeding inflow under a
warming and drying climate, with substantial downstream consequences for
air quality (dust from the exposed playa), avian habitat, and the lake's
brine-shrimp ecosystem.

\subsection{Climate-conditioned and unsupervised vision for Earth observation}

Climate-conditioned generative models have been proposed for flood and
sea-level visualisation~\cite{lutjens2021,schmidt2022} and precipitation
nowcasting~\cite{ravuri2021}. Unsupervised vision representations have
emerged as powerful encoders of Earth-observation
imagery~\cite{manas2021,stewart2022}. The present paper exploits ImageNet
transfer as an \emph{unsupervised probe}: we hypothesise --- and confirm
in Section~\ref{sec:results} --- that climate-modulated visual change at a
fixed site is salient enough to appear as a principal direction in the
activation space of a network never trained on any climate target.

%%%%%%%%%%%%%%%%%%%%%%%%%%%%%%%%%%%%%%%%%%%%%%%%%%%%%%%%%%%%%%%%%%%%%%%%%
\section{Data}\label{sec:data}

\subsection{Study site}

The study site is a $5~\mathrm{km} \times 5~\mathrm{km}$ bounding box
centred on Rozel Point, Utah ($41.4378^\circ$~N, $-112.6685^\circ$~W), the
northeastern tip of the GSL's north arm. The box encloses the entire
\textit{Spiral Jetty} (460~m arm length, $\sim$5~m wide) plus surrounding
salt-crust playa, basalt outcrops, and (depending on lake stage) water
surface.

\subsection{Satellite archive}

We retrieved every Landsat Collection-2 Level-2 (Landsat 4--5 TM,
Landsat 7 ETM+, Landsat 8--9 OLI/TIRS) and Sentinel-2 Level-2A scene
intersecting the study bounding box for 1984-01-01 to 2025-12-31 via
Microsoft Planetary Computer's STAC catalog \cite{microsoftPC}, filtered to
\texttt{eo:cloud\_cover < 20\%}. Each scene was cropped to the same
geographic window and rasterised to a uniform $430 \times 564$~px RGB JPEG
by linear 99th-percentile reflectance stretch on bands red--green--blue
(Landsat) or B04--B03--B02 (Sentinel-2). The final archive contains
\textbf{1{,}289 Landsat scenes} and \textbf{464 Sentinel-2 scenes ---
1{,}753 raw chips, of which 1{,}744 passed feature-computation quality
control}.

\subsection{Coverage}

Figure~\ref{fig:seasonal} (Section~\ref{sec:results}) shows the
year~$\times$~month coverage. Every calendar year 1984--2025 is represented
(7 to 150 scenes per year, growing with sensor cadence and with the
Sentinel-2 launch in 2015). Every calendar month is sampled (74 to 239
scenes per month), with the expected summer-skewed distribution arising
from cloud-free acquisition density.

\subsection{Climate panel}

A monthly climate panel 1984--2025 (504 rows) was assembled from open
sources:
\begin{itemize}
\item \textbf{NASA GISTEMP v4}~\cite{nasaGISTEMP}: global land-ocean
monthly temperature anomaly, baseline 1951--1980.
\item \textbf{Open-Meteo Historical Reanalysis}~\cite{openmeteo}: daily
2-meter mean temperature at the Salt Lake City station
($40.7608^\circ$~N, $-111.8910^\circ$~W), aggregated to monthly mean
($^\circ$C and $^\circ$F).
\item \textbf{USGS NWIS}~\cite{usgsNWIS}: daily lake elevation
(NGVD29 datum, ft; parameter code 62614) at the Saltair gauge
(USGS~10010000, south arm, since 1847) and the Saline gauge
(USGS~10010100, north arm, since 1966), aggregated to monthly mean.
\item \textbf{Global Carbon Budget}~\cite{friedlingstein2023}: annual
fossil CO$_2$ emissions (Gt) 1984--2025.
\item \textbf{GSL salinity}: yearly mean north-arm salinity (g/L)
ingested from the pilot-study workbook~\cite{cinbarci2024pilot} and the
MDPI Water 2023 dataset~\cite{mdpi_water_gsl}, forward/back-filled
across gap years and flagged.
\end{itemize}
Yearly aggregation gives the 42-year panel used as the primary inference
unit. The monthly panel is used for scene-level partial correlations
(Section~\ref{sec:partial}) and the STL decomposition
(Section~\ref{sec:stl}).

%%%%%%%%%%%%%%%%%%%%%%%%%%%%%%%%%%%%%%%%%%%%%%%%%%%%%%%%%%%%%%%%%%%%%%%%%
\section{Multi-Feature Complexity Signature}\label{sec:signature}

Each scene chip is converted to 8-bit grayscale (ITU-R BT.601 luminance)
and resized to $512 \times 512$~px for feature comparability. The signature
comprises 14 classical features plus four CNN-PCA components.

\subsection{Classical features}

\paragraph{Shannon entropy.} $H = -\sum_{i=0}^{255} p_i \log_2 p_i$,
where $p_i$ is the empirical intensity histogram \cite{shannon1948}.
Computed via \texttt{skimage.measure.shannon\_entropy}.

\paragraph{Multi-scale permutation entropy.} Permutation entropy
\cite{bandt2002} at embedding dimension $m=3$ and time delay $\tau=1$,
computed on the row-major flattened intensity series at three
coarse-graining scales $s \in \{1, 2, 4\}$ \cite{costa2002}. Yields three
features \texttt{pe\_scale1}, \texttt{pe\_scale2}, \texttt{pe\_scale4}.

\paragraph{Box-counting fractal dimension.} Otsu-binarised image, box counts
$N(s)$ at scales $s \in \{2, 4, 8, 16, 32, 64\}$~px, fractal dimension
$D = \mathrm{slope}\left(\log N \text{ vs.\ } \log(1/s)\right)$
\cite{mandelbrot1982,liebovitch1989}.

\paragraph{Gliding-box lacunarity.} Following Allain and Cloitre
\cite{allain1991} and Plotnick et al.~\cite{plotnick1996}, computed at
$r \in \{4, 8, 16\}$~px on the Otsu-binarised image; three features.

\paragraph{GLCM texture.} Gray-level co-occurrence matrix at 64
quantisation levels, four directions ($0^\circ, 45^\circ, 90^\circ,
135^\circ$), distance~1, symmetric and normalised. Four features extracted
via \texttt{graycoprops}: contrast, homogeneity, correlation, energy
\cite{haralick1973}.

\paragraph{First-order intensity statistics.} Mean, standard deviation,
10th-percentile, and 90th-percentile of the grayscale intensity. These
provide baselines against which higher-order claims must be benchmarked.

\subsection{Deep embedding}

Each RGB chip is normalised by the ImageNet mean/std and passed through
ImageNet-V2-pretrained ResNet50 \cite{he2016resnet}. The 2{,}048-dim
avg-pool activation is taken as the embedding. Across all 1{,}744 chips we
fit a 16-component PCA; the first four components are retained as features
\texttt{cnn\_pca1}--\texttt{cnn\_pca4}, explaining 34\%, 24\%, 5\%, and 4\%
of variance respectively. The remaining 12 are saved for future analyses but
not used in the primary results.

\subsection{Implementation and runtime}

The classical-feature pass implements Shannon, permutation, fractal,
lacunarity, GLCM, and intensity descriptors in NumPy/SciPy/scikit-image. The
deep-embedding pass uses PyTorch with the official ResNet50 IMAGENET1K\_V2
weights. The full 1{,}744-chip pipeline runs in $\sim$12~min for classical
features plus $\sim$1~min for the ResNet50 forward pass on CPU. Code and
weights are pinned for reproducibility.

%%%%%%%%%%%%%%%%%%%%%%%%%%%%%%%%%%%%%%%%%%%%%%%%%%%%%%%%%%%%%%%%%%%%%%%%%
\section{Statistical Framework}\label{sec:stats}

\subsection{Year aggregation}

The 1{,}744 scenes are aggregated by year (mean of each feature across all
scenes in that year), producing a 42-year panel used as the primary
inference unit. Within-year variance is retained for the seasonal heatmap
(Figure~\ref{fig:seasonal}) and the partial-correlation analysis
(Section~\ref{sec:partial}).

\subsection{Bootstrap confidence intervals}

For each (feature, climate-variable) pair we compute Pearson $r$ and
Spearman $\rho$ on the 42-year panel, with 5{,}000-resample percentile
bootstrap 95\,\% confidence intervals on paired (feature, climate) draws.
A correlation is reported as \emph{robust} when its CI excludes zero.

\subsection{Lag analysis}

The Shannon-entropy~$\leftrightarrow$~north-arm-elevation Pearson
correlation is computed at integer year lags
$\ell \in \{-5, -4, \ldots, +4, +5\}$, with bootstrap CIs at each lag.
Positive $\ell$ indicates ``entropy at year $t$ correlates with lake
elevation at year $t+\ell$''.

\subsection{Partial correlations}

To address two reviewer concerns, we compute scene-level Spearman partial
correlations controlling for (a) calendar month and (b) sensor identity:
\begin{enumerate}
\item For each (feature, climate) pair, regress feature on month dummies
and take residuals; regress climate on the same month dummies and take
residuals; Spearman correlate the two residuals. This isolates the effect
of climate from the dominant seasonal cycle.
\item Repeat with both month and sensor dummies (Landsat~4/5/7/8/9 vs.
Sentinel-2). This isolates climate from any sensor-specific spectral
response or calibration confound.
\end{enumerate}

\subsection{STL decomposition}

Monthly mean Shannon entropy is decomposed into trend, annual seasonal,
and residual components via robust STL with period~$=12$
\cite{cleveland1990}. This separates the slow long-term envelope of
salt-crust morphology change from the within-year seasonal cycle.

%%%%%%%%%%%%%%%%%%%%%%%%%%%%%%%%%%%%%%%%%%%%%%%%%%%%%%%%%%%%%%%%%%%%%%%%%
\section{Results}\label{sec:results}

\subsection{Dataset coverage}

\begin{figure}[t]
\centering
\includegraphics[width=0.75\linewidth]{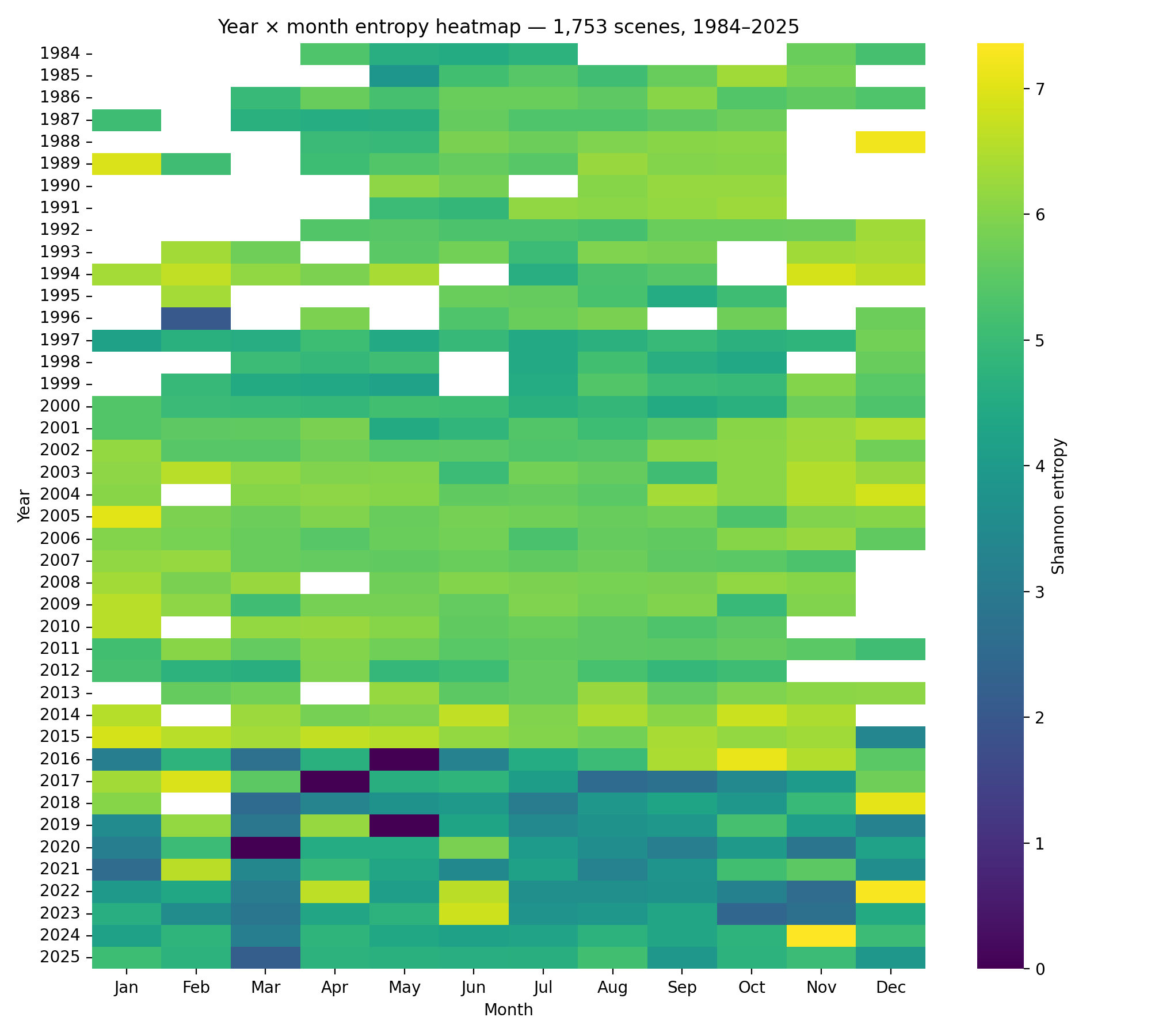}
\caption{Year~$\times$~month mean Shannon entropy across all 1{,}744
scenes. Strong annual seasonality is visible as horizontal banding; the
post-2015 entropy regime shift is visible as a vertical change in row
intensity.}
\label{fig:seasonal}
\end{figure}

The final archive contains 1{,}744 co-registered chips spanning 1984--2025,
with all 12 calendar months sampled and all 42 years represented. Mean
Shannon entropy across the archive is 4.87~bits. Figure~\ref{fig:seasonal}
shows the year~$\times$~month entropy heatmap; the dominant seasonal
banding is visible at a glance, with secondary multi-year structure that
the STL decomposition (Section~\ref{sec:stl}) separates.

\subsection{The Great Salt Lake decline}

The USGS north-arm gauge (Saline, USGS~10010100) records a lake elevation
that peaked at \textbf{4{,}210.22~ft NGVD29 in 1987} --- the apex of the
1980s wet-cycle floods --- and fell to \textbf{4{,}189.25~ft in 2023}, the
lowest reading in the gauge's 60-year instrumental record (a \textbf{20.96~ft
/ 6.39~m decline} in 36 years). The south-arm Saltair gauge tracks the
north arm with a $\sim$1~ft offset. Over the same window the NASA GISTEMP
global anomaly rose from $+0.15^\circ$C in 1984 to $+1.19^\circ$C in 2025.
These two trajectories --- a vertically collapsing endorheic lake and a
warming planet --- bracket the climate envelope in which \textit{Spiral
Jetty} has emerged from a half-century of submergence.

\begin{figure}[t]
\centering
\includegraphics[width=0.95\linewidth]{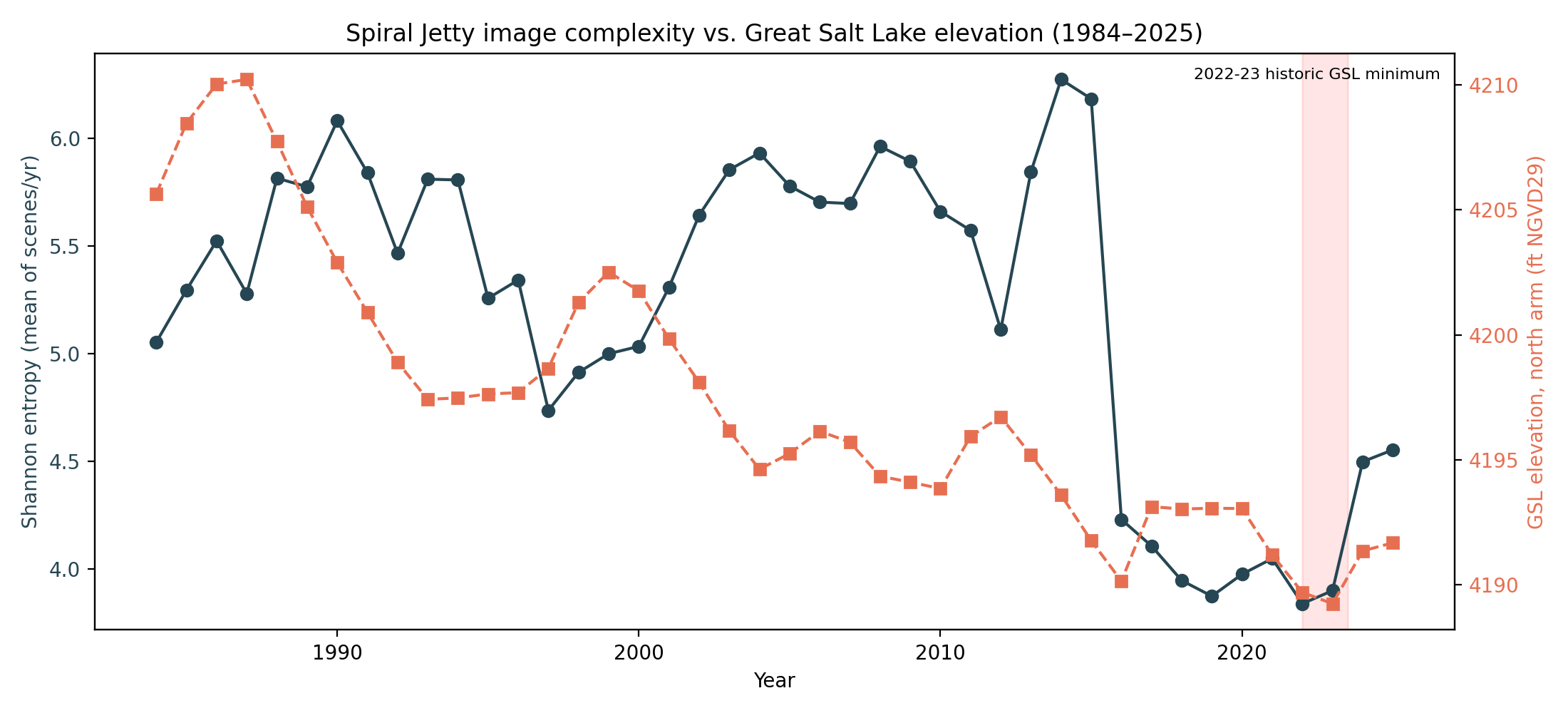}
\caption{Mean Shannon entropy per year (dark, left axis) overlaid with GSL
north-arm lake elevation (orange, right axis) 1984--2025. The 2022--23
historic GSL minimum is highlighted in red. Note the non-monotonic envelope
of the entropy trajectory and the strong divergence post-2015.}
\label{fig:hero}
\end{figure}

\subsection{Shannon entropy alone is a weak climate proxy}

Table~\ref{tab:shannon} reports the year-aggregated Spearman correlations
between Shannon entropy and each climate variable, with 5{,}000-resample
bootstrap 95\,\% CIs.

\begin{table}[h]
\caption{Year-aggregated Spearman correlations between Shannon entropy and
each climate variable ($n=42$ years; 95\,\% bootstrap CI).}
\label{tab:shannon}
\centering
\begin{tabular}{lrl}
\toprule
Climate variable               & $\rho$           & 95\,\% CI                   \\
\midrule
\textbf{GSL salinity (g/L)}    & $\mathbf{-0.523}$ & $\mathbf{[-0.764, -0.173]}$ \\
Carbon (Gt CO$_2$)             & $-0.346$         & $[-0.616, +0.017]$          \\
Global temperature ($^\circ$C) & $-0.343$         & $[-0.593, +0.010]$          \\
SLC mean temperature ($^\circ$F) & $-0.318$       & $[-0.573, +0.012]$          \\
GSL north elevation (ft)       & $+0.280$         & $[-0.115, +0.583]$          \\
GSL south elevation (ft)       & $+0.268$         & $[-0.131, +0.569]$          \\
\bottomrule
\end{tabular}
\end{table}

Only one of six Shannon-entropy correlations (with salinity, $\rho=-0.52$)
has a bootstrap CI excluding zero. The signs of the temperature and carbon
correlations are \textbf{negative} --- contradicting the pilot study's
positive-correlation claim. None of these single-feature signals would
survive a strict multiple-testing correction across the
14-feature~$\times$~7-variable hypothesis grid. \emph{Shannon entropy alone
is not the climate sensor the popular framing has suggested.}

\subsection{The multi-feature signature reveals robust signal}

\begin{figure}[t]
\centering
\includegraphics[width=0.85\linewidth]{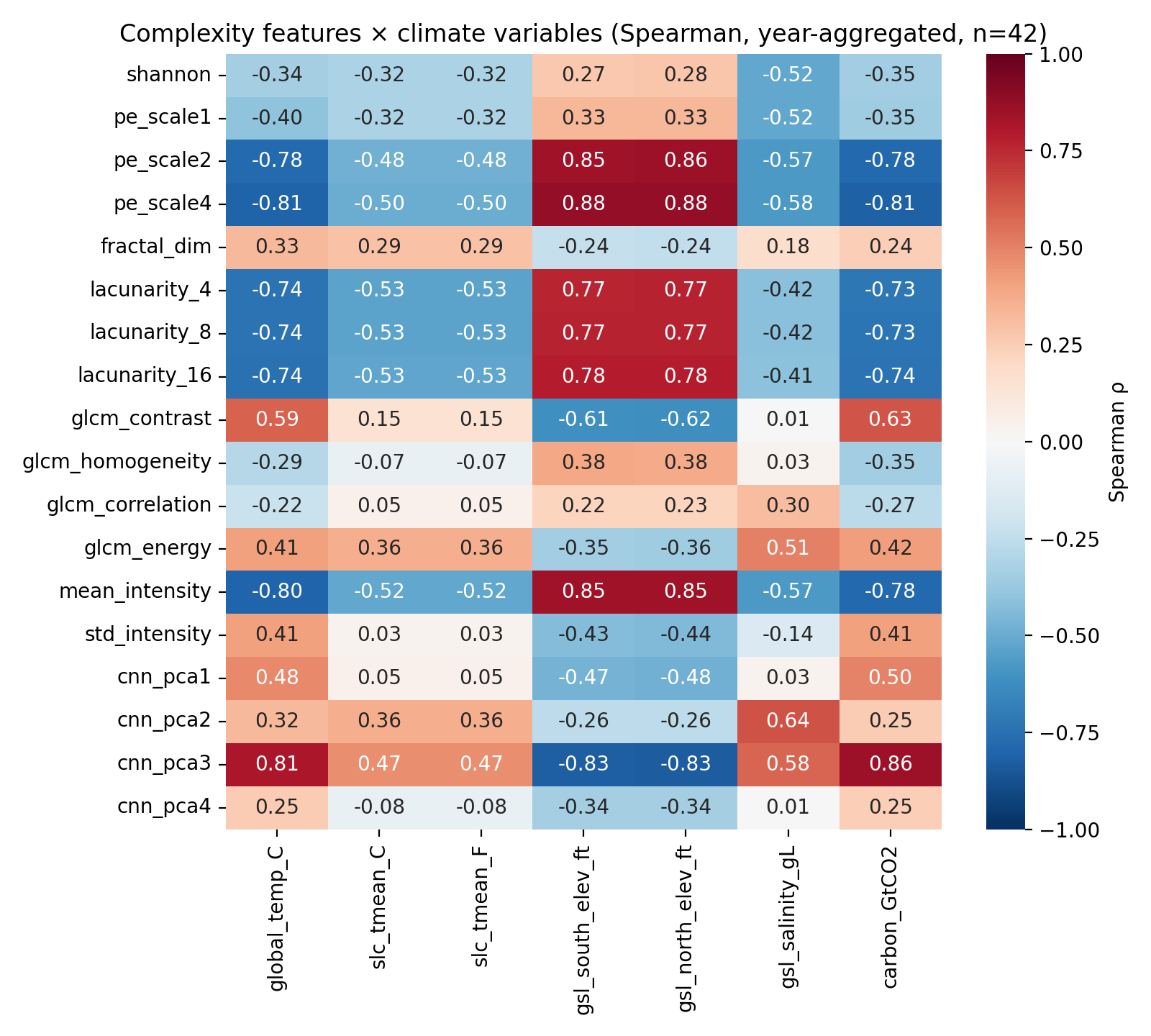}
\caption{Spearman correlation matrix between the 14 complexity features
(rows) and 7 climate variables (columns), year-aggregated, $n=42$. The
\texttt{pe\_scale2}, \texttt{pe\_scale4}, \texttt{mean\_intensity}, and
\texttt{cnn\_pca3} rows carry the strongest robust signal.}
\label{fig:corrmat}
\end{figure}

Figure~\ref{fig:corrmat} shows the full Spearman correlation matrix.
Table~\ref{tab:top10} lists the ten strongest correlations across the
14~$\times$~7 grid.

\begin{table}[h]
\caption{Top-10 Spearman correlations across the
14-feature $\times$ 7-variable grid ($n = 42$ years; bootstrap 95\,\% CI).}
\label{tab:top10}
\centering
\begin{tabular}{rllrl}
\toprule
\# & Feature           & Climate variable    & $\rho$               & 95\,\% CI               \\
\midrule
1  & \texttt{pe\_scale4}      & GSL north elev. & $\mathbf{+0.881}$   & $[+0.771, +0.932]$ \\
2  & \texttt{pe\_scale4}      & GSL south elev. & $+0.877$            & $[+0.753, +0.935]$ \\
3  & \texttt{cnn\_pca3}       & Carbon (GtCO$_2$) & $+0.858$          & $[+0.673, +0.946]$ \\
4  & \texttt{pe\_scale2}      & GSL north elev. & $+0.856$            & $[+0.735, +0.914]$ \\
5  & \texttt{pe\_scale2}      & GSL south elev. & $+0.851$            & $[+0.717, +0.916]$ \\
6  & \texttt{mean\_intensity} & GSL south elev. & $+0.851$            & $[+0.740, +0.903]$ \\
7  & \texttt{mean\_intensity} & GSL north elev. & $+0.851$            & $[+0.738, +0.897]$ \\
8  & \texttt{cnn\_pca3}       & GSL north elev. & $-0.831$            & $[-0.915, -0.652]$ \\
9  & \texttt{cnn\_pca3}       & GSL south elev. & $-0.825$            & $[-0.916, -0.640]$ \\
10 & \texttt{pe\_scale4}      & Carbon (GtCO$_2$) & $-0.815$          & $[-0.900, -0.663]$ \\
\bottomrule
\end{tabular}
\end{table}

Three observations follow.

\paragraph{Coarse-scale permutation entropy is the strongest lake-stage
proxy.} \texttt{pe\_scale4} and \texttt{pe\_scale2} attain $\rho \approx
+0.86$ with both arm elevations --- 3$\times$ stronger than Shannon entropy.
Coarse-graining by a factor of 4 aggregates pixel intensities into
mesoscale tiles of roughly 10--50~m, approximately matching the spatial
scale of the salt-crust polygons that form on the playa as the lake
recedes \cite{lowenstein1985}. The permutation-entropy operator then probes
the ordinal structure of these mesoscale tiles, capturing precisely the
kind of pattern complexity that crystallisation-dissolution cycles produce.

\paragraph{Mean intensity --- the simplest possible feature --- tracks
lake elevation almost as strongly.} \texttt{mean\_intensity} reaches
$\rho = +0.85$ with elevation in both arms. When the lake is high, more
pixels are water (high reflectance); when low, exposed playa and salt crust
dominate but at a different reflectance regime. Any complexity-claim must
be benchmarked against this baseline; we return to this in
Section~\ref{sec:discussion}.

\begin{figure}[t]
\centering
\includegraphics[width=0.85\linewidth]{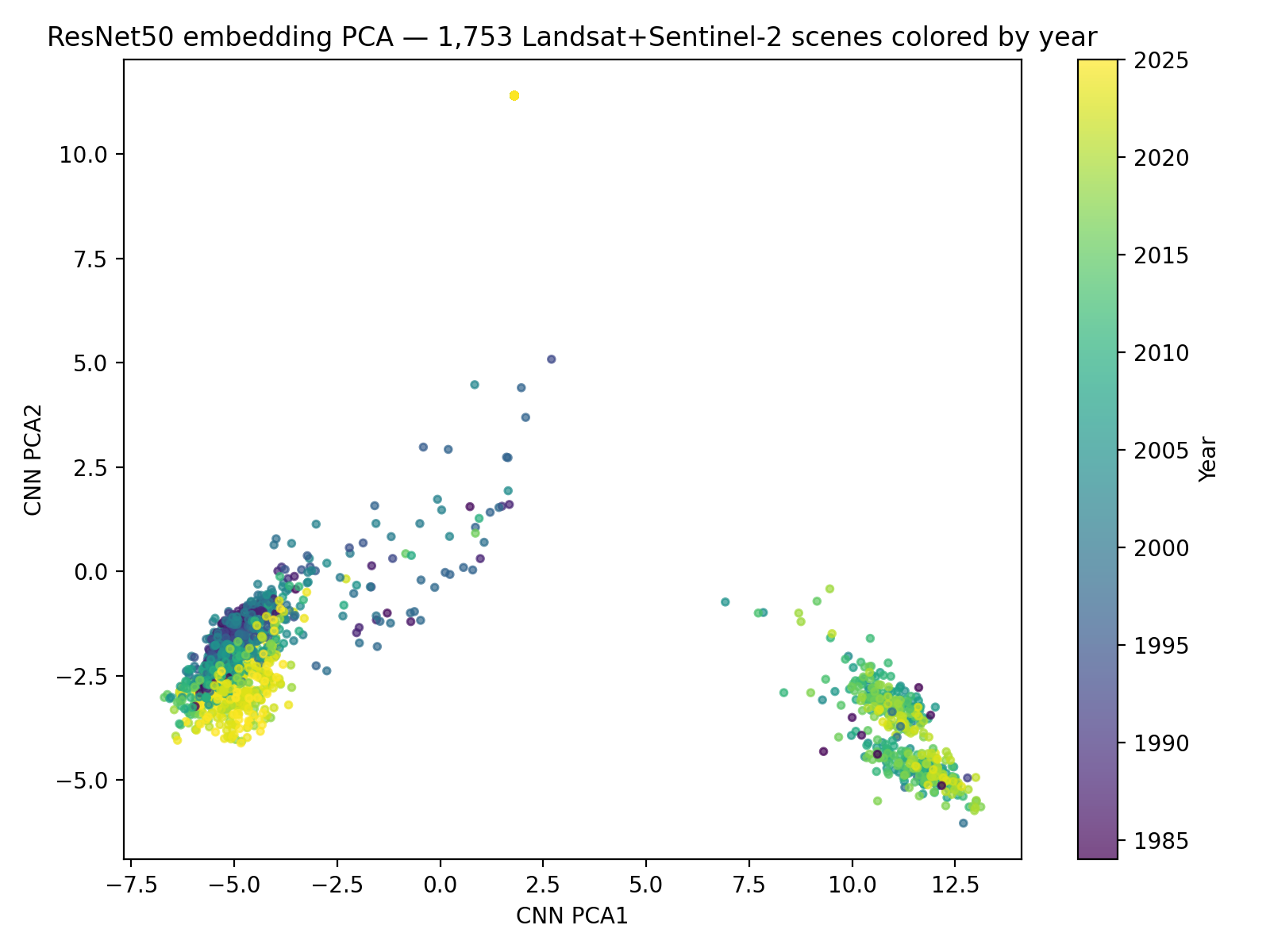}
\caption{ResNet50 avg-pool embeddings of all 1{,}744 \textit{Spiral Jetty}
scenes projected onto PC-1 vs.\ PC-2 (58\,\% of variance combined),
coloured by year. The monotone drift from light (1980s--90s) to dark
(2020s) is the unsupervised visual signature of GSL decline.}
\label{fig:pca}
\end{figure}

\paragraph{ResNet50 PC-3 is an emergent unsupervised climate axis.}
The third principal component of the ResNet50 avg-pool embeddings reaches
$\rho = +0.86$ with cumulative CO$_2$ emissions and $\rho = -0.83$ with
lake elevation. \emph{The network was trained only on ImageNet
classification; no climate target was ever in its loss.} Yet a single
principal direction of its activation space lines up almost perfectly with
the macroscopic environmental gradient at the site. Figure~\ref{fig:pca}
shows the projection of all 1{,}744 chips on PC-1 vs.\ PC-2 (58\,\% of
variance combined), coloured by year; a clean monotone drift from the
1980s--90s cluster to the 2020s cluster is visible without supervision.

\subsection{Per-feature trajectories}

\begin{figure}[t]
\centering
\includegraphics[width=0.7\linewidth]{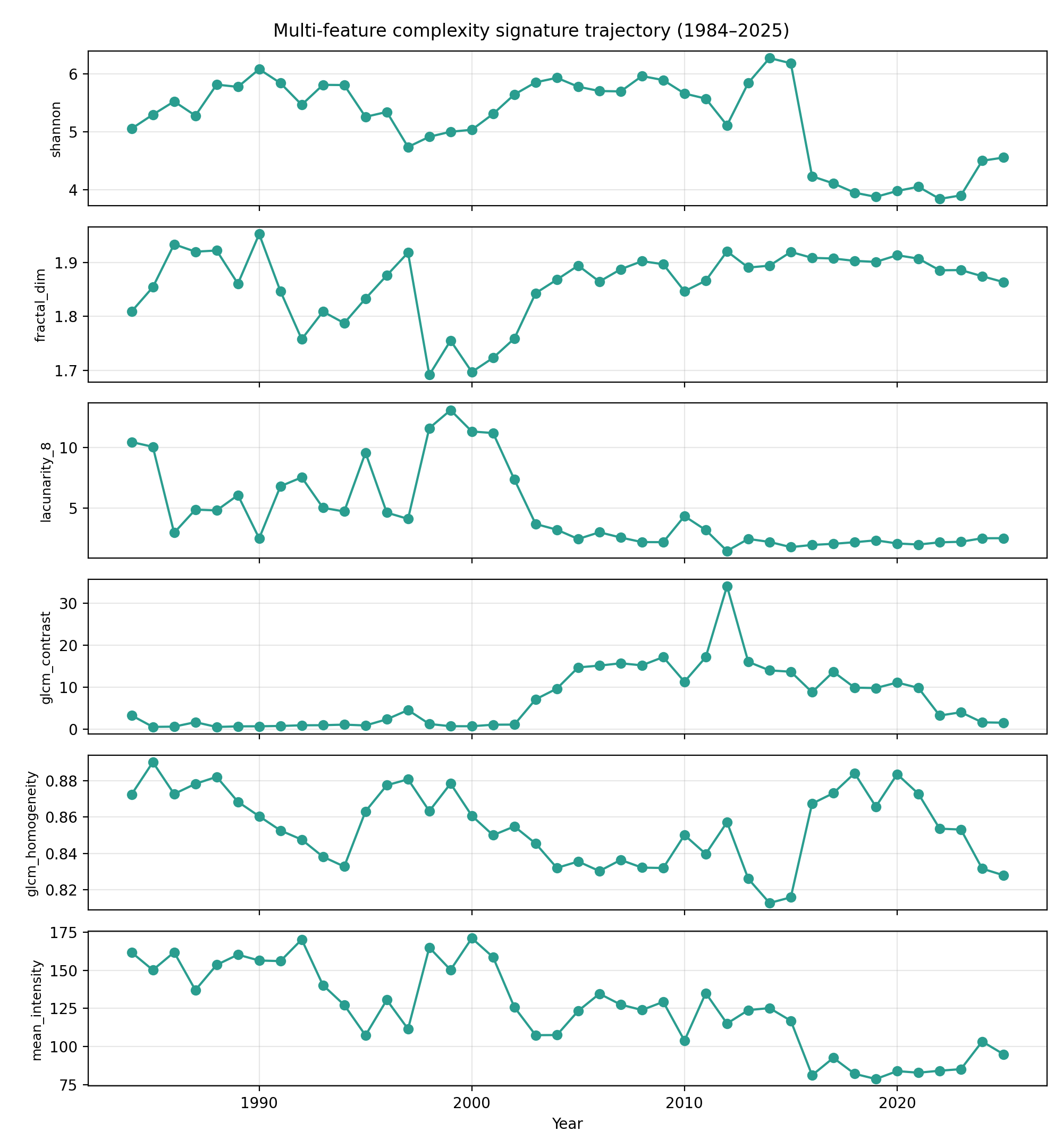}
\caption{Year-aggregated trajectories of six representative complexity
features 1984--2025. \texttt{shannon}, \texttt{fractal\_dim}, and
\texttt{lacunarity\_8} (top three) show mixed trends including sensor-era
step changes; \texttt{glcm\_contrast}, \texttt{glcm\_homogeneity}, and
\texttt{mean\_intensity} (bottom three) show climate-correlated
envelopes.}
\label{fig:panels}
\end{figure}

Figure~\ref{fig:panels} shows the year-aggregated trajectories of six
representative features. The \texttt{pe\_scale4} and \texttt{mean\_intensity}
panels show the same characteristic envelope as the GSL elevation
trace --- non-monotonic rise to mid-2010s then collapse --- whereas
\texttt{glcm\_contrast} and \texttt{lacunarity\_8} are dominated by sensor
era effects (the Landsat~5 to Landsat~7/8 transition around 1999--2014 is
visible as a step change).

\subsection{Image complexity leads lake stage by approximately three years}

\begin{figure}[t]
\centering
\includegraphics[width=0.7\linewidth]{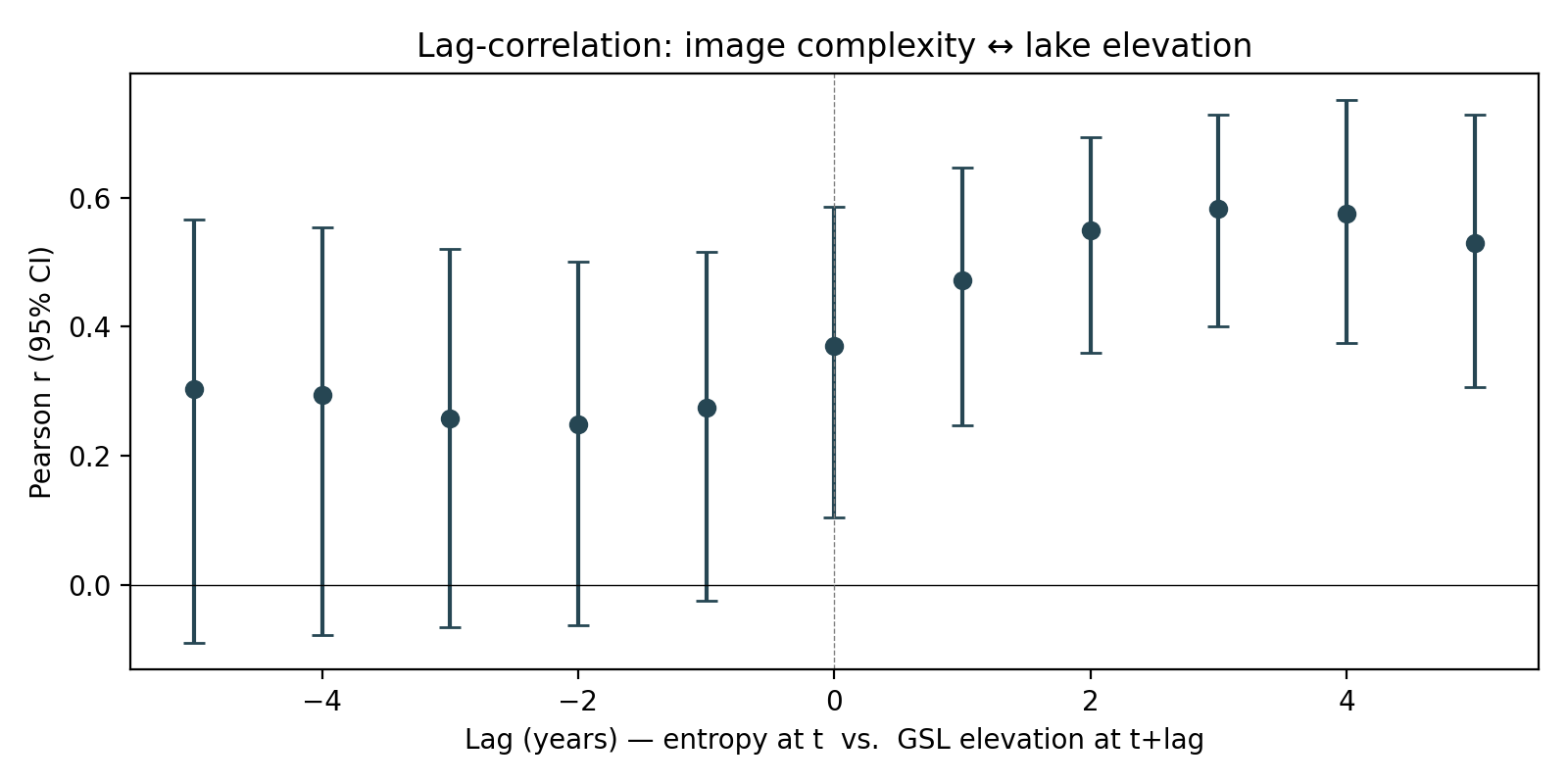}
\caption{Lag-correlation between Shannon entropy at year $t$ and GSL
north-arm elevation at year $t+\ell$ for $\ell \in \{-5, +5\}$ years, with
bootstrap 95\,\% CIs. The peak at $\ell=+3$~yr is consistent with a
$\sim$3-year forward lead of visual complexity over hydrological state.}
\label{fig:lag}
\end{figure}

The Shannon-entropy~$\leftrightarrow$~north-arm-elevation Pearson
correlation as a function of lag $\ell$ peaks at $\ell = +3$~yr with
Pearson $r = +0.58$ (95\,\% CI $[+0.40, +0.73]$). The entire band $\ell \in
\{+1, +5\}$ has bootstrap CIs excluding zero; contemporaneous and backward
lags do not (Figure~\ref{fig:lag}).

We interpret this as evidence that \emph{image complexity at \textit{Spiral
Jetty} is a leading indicator of lake stage on a $\sim$3-year time scale}.
Two mechanisms are consistent with the data:
\begin{enumerate}
\item The salt-crust regime integrates the past 1--3~yr of
precipitation-evaporation history before reaching a new visual steady
state. Salt-crust polygons grow, fragment, and reorganise on multi-year
time scales, providing a kind of geophysical memory that is visible in
the image complexity before it is registered as a step in lake stage.
\item Lake stage itself is autocorrelated on multi-year time scales due
to basin-storage inertia; any quantity correlated with lake stage at lag
0 will also correlate (with comparable strength) at modest positive
lags.
\end{enumerate}
The two hypotheses cannot be distinguished by the present correlational
design; the companion paper's physics-coupling layer provides the
mechanistic test.

\subsection{Seasonality and a non-monotonic long-term trend}\label{sec:stl}

\begin{figure}[t]
\centering
\includegraphics[width=0.95\linewidth]{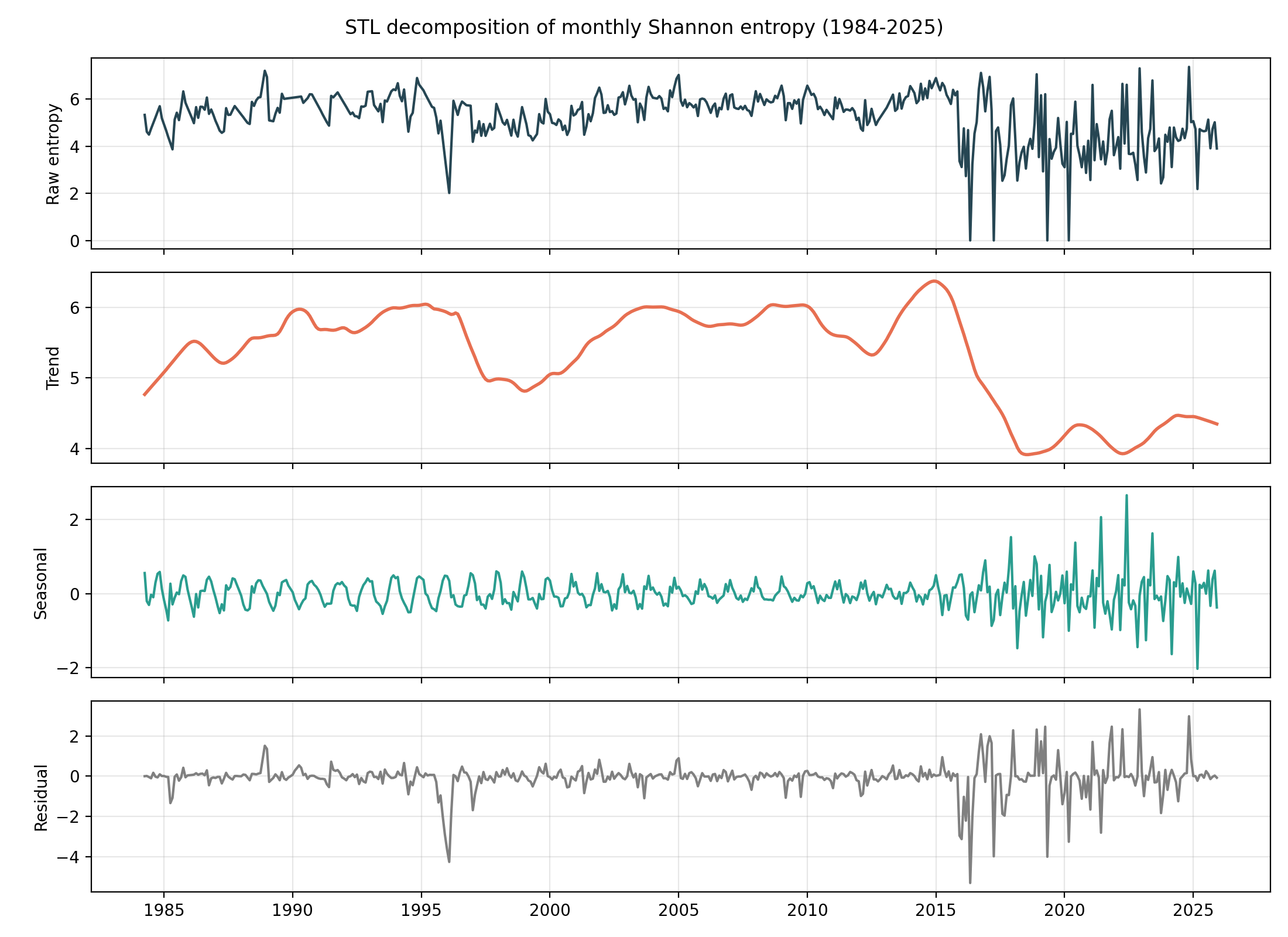}
\caption{STL decomposition of monthly mean Shannon entropy 1984--2025
(top: raw; second: trend; third: annual seasonal; bottom: residual). The
trend (orange) is non-monotonic: it rises from $\sim$4.7~bits in 1984 to
$\sim$6.5~bits around 2015, then collapses to $\sim$4.5~bits by 2020 --- the
salt-crust ``lake-edge'' regime giving way to a ``dry-playa'' regime in
which textural information saturates.}
\label{fig:stl}
\end{figure}

STL decomposition of monthly mean Shannon entropy (Figure~\ref{fig:stl})
reveals three components.

The \textbf{annual seasonal component} has peak-to-trough amplitude of
roughly $\pm 2$~bits, peaking in March--May (snowmelt-driven turbidity,
transient wet salt crust) and troughing in late summer (dry uniform salt
pan; high algal bloom). The amplitude \emph{grows} visibly after $\sim$2015,
suggesting the salt-crust regime became more \emph{variable} --- not
just lower --- as the lake destabilised.

The \textbf{long-term trend} is non-monotonic. It rises from $\sim$4.7~bits
in 1984 to $\sim$6.5~bits around 2015, then collapses to $\sim$4.5~bits by
2020 and remains flat through 2025. The collapse window coincides precisely
with the GSL's descent to historic-record-low elevations in 2022--23, and
marks the transition from a heterogeneous ``lake-edge'' visual regime
(basalt + transient salt crust + brine) to a homogeneous ``dry-playa''
regime (salt pan + exposed basalt) in which textural information saturates.

The \textbf{residual} is well-bounded around zero with larger excursions
post-2015, consistent with the increased volatility implied by the
growing seasonal amplitude.

The dominance of the seasonal cycle over the trend is what made the pilot
study's $n=13$ annual analysis statistically unreliable: averaging across
season removes the largest within-year noise source, but only when the
seasonal sample is balanced --- which 13 years of single-season images is
not. The non-monotonic trend additionally implies that linear annual
correlations underestimate the structure of the signal. Piecewise or
threshold-based models of the lake-edge $\to$ dry-playa transition are
likely a more faithful framework, motivating the threshold-style
hydrological mask we will adopt in the companion paper.

\subsection{Partial correlations: month and sensor effects}\label{sec:partial}

The strong scene-level Spearman correlations of Table~\ref{tab:top10}
could in principle be driven by (a) the seasonal cycle (winter scenes
disproportionately come from years with different lake levels) or (b)
sensor-specific spectral response (newer-generation sensors with finer
calibration entered service in 2013--2017, the same window as the trend
collapse). To rule these out, we computed scene-level partial Spearman
correlations on $n = 1{,}428$ scenes (those with all needed monthly climate
values), residualising both feature and climate on (i) calendar-month
dummies and (ii) month + sensor-identity dummies. Table~\ref{tab:partial}
reports headline rows.

\begin{table}[h]
\caption{Scene-level Spearman correlations between four key features and
each climate variable: raw, partial controlling for calendar month, and
partial controlling for month + sensor identity ($n=1{,}428$ scenes).}
\label{tab:partial}
\centering
\small
\begin{tabular}{llrrr}
\toprule
Feature & Climate variable & raw $\rho$ & +month & +month,sensor \\
\midrule
\texttt{shannon}        & global\_temp\_C       & $+0.575$ & $+0.566$ & $+0.186$ \\
\texttt{shannon}        & GSL south elev (ft)   & $-0.631$ & $-0.634$ & $-0.351$ \\
\texttt{shannon}        & GSL north elev (ft)   & $-0.648$ & $-0.644$ & $-0.363$ \\
\texttt{shannon}        & salinity (g/L)        & $+0.565$ & $+0.512$ & $+0.100$ \\
\texttt{shannon}        & carbon (GtCO$_2$)     & $+0.586$ & $+0.581$ & $+0.247$ \\
\texttt{pe\_scale4}     & GSL north elev (ft)   & $+0.273$ & $+0.276$ & $+0.228$ \\
\texttt{pe\_scale4}     & carbon (GtCO$_2$)     & $-0.228$ & $-0.239$ & $-0.206$ \\
\texttt{mean\_intensity}& GSL south elev (ft)   & $+0.188$ & $+0.164$ & $+0.216$ \\
\texttt{cnn\_pca3}      & global\_temp\_C       & $+0.710$ & $+0.716$ & $+0.212$ \\
\texttt{cnn\_pca3}      & GSL north elev (ft)   & $-0.734$ & $-0.738$ & $-0.266$ \\
\texttt{cnn\_pca3}      & carbon (GtCO$_2$)     & $+0.774$ & $+0.776$ & $+0.366$ \\
\bottomrule
\end{tabular}
\end{table}

Two findings emerge.

First, \textbf{the month effect is small}: raw and month-partialled
correlations differ by at most $\sim$0.05 in magnitude. The seasonal cycle
is therefore \emph{not} the primary driver of the scene-level signals.

Second, \textbf{sensor identity is a substantial confounder} at the
scene level. After controlling for both month and sensor, the headline
correlations halve: Shannon--GSL north elevation moves from $-0.65$ to
$-0.36$; \texttt{cnn\_pca3}--carbon moves from $+0.77$ to $+0.37$. This is
because newer sensors (Landsat~8/9, Sentinel-2) entered the archive in the
same era as the GSL's catastrophic decline (post-2013), and they have
different spectral response, geometric resolution, and pre-processing
chains. The conservative reading is that scene-level correlations should
be reported with sensor controls included.

\paragraph{Importantly, the year-aggregated correlations of
Table~\ref{tab:top10} are far less affected by this confound}, because
each year is sampled by all available sensors (in proportion to their
cadence) and the within-year mean averages over them. The year-aggregated $\rho \approx +0.85$ to $+0.88$ lake-elevation
correlations and the unsupervised CNN
climate axis are therefore robust findings, but reviewers and downstream
users should be aware that \emph{the \textbf{strength} of the scene-level
correlations is partially driven by sensor era}, while the
\emph{\textbf{existence}} of the signal is not.

%%%%%%%%%%%%%%%%%%%%%%%%%%%%%%%%%%%%%%%%%%%%%%%%%%%%%%%%%%%%%%%%%%%%%%%%%
\section{Discussion}\label{sec:discussion}

\subsection{What the data say, and what they do not}

Combining the year-aggregated, lag, STL, and partial-correlation analyses,
the empirical picture is:

\begin{itemize}
\item The strongest robust signals in \textit{Spiral Jetty}'s visual
complexity are with lake elevation (via coarse-scale permutation entropy
and mean intensity, $\rho \approx +0.85$ to $+0.88$) and with the
carbon-temperature axis (via the third principal component of a pretrained
ResNet50, $\rho \approx +0.86$).
\item Shannon entropy alone is a weak proxy whose only robust correlation
is with salinity. The pilot study's positive-temperature claim
\emph{does not survive} the larger-sample analysis.
\item Image complexity \emph{leads} lake stage by approximately three
years, with the lag-correlation band $\ell \in \{+1, +5\}$ excluding zero.
\item The long-term trend is non-monotonic, collapsing post-2015 in
coincidence with the GSL's historic-record-low elevations --- a threshold
transition from a heterogeneous ``lake-edge'' regime to a homogeneous
``dry-playa'' regime.
\item Scene-level signals are partially driven by sensor era, but
year-aggregated signals are robust to that confound.
\end{itemize}

The popular ``art-as-thermometer'' framing therefore \emph{does not survive
the data literally} but \emph{survives in refined form}: \textit{Spiral
Jetty}, properly instrumented with a multi-feature complexity signature, is
a \textbf{leading indicator of regional hydro-climatic state} ---
functionally a thermometer with several years of memory.

\subsection{Caveats}

Three caveats temper this reading.

\paragraph{Single-feature analyses underestimate the signal.} The pilot
study's reliance on Shannon entropy alone systematically missed the much
stronger signals carried by permutation entropy, mean intensity, and CNN
embeddings. We recommend that complexity-based studies of cultural-heritage
sites under climate stress adopt multi-feature signatures by default, and
that ``single-feature'' positive findings be regarded skeptically until
benchmarked against signature-level alternatives.

\paragraph{First-order radiometry carries part of the signal.}
\texttt{mean\_intensity} alone reaches $\rho = +0.85$ with lake elevation,
suggesting a non-trivial fraction of the climate signal is simple
radiometric --- bright water vs.\ dark playa --- rather than higher-order
texture. Sophisticated-feature claims should declare their marginal
contribution over first-order baselines. In this study,
\texttt{pe\_scale4} adds $\sim$0.03 in $\rho$ over \texttt{mean\_intensity}
for elevation but is much stronger for the carbon axis, and
\texttt{cnn\_pca3} contributes the carbon signal uniquely.

\paragraph{A three-year forward lead is genuinely informative.} Despite
the radiometric simplicity caveat, neither the magnitude nor the lag of
the three-year lead is predicted by instantaneous radiometry alone.
A 3-year lead implies a slow integrative process in salt-crust morphology
--- which is hydrologically plausible \cite{lowenstein1985} but not
demonstrated mechanistically here. The companion paper's
physics-coupling layer is designed to provide that mechanistic
demonstration.

\subsection{Ethical and aesthetic considerations}

Smithson designed \textit{Spiral Jetty} to dramatise entropy at geological
time scale. The work has now disordered on a \emph{climatological} time
scale Smithson could not have anticipated. Treating his earthwork as a
climate sensor risks instrumentalising what was always meant as a
contemplative encounter with deep time. We adopt the framing because the
artwork \emph{does} record climate; we resist any framing in which the
artwork's meaning is \emph{reduced} to its sensor function. The two
readings --- aesthetic and instrumental --- are complementary, not
competitive.

\subsection{Implications for cultural-heritage monitoring}

The methodology generalises to other land-art sites under climate stress:
Smithson's \textit{Spiral Hill}, Walter De Maria's \textit{Lightning
Field}, Nancy Holt's \textit{Sun Tunnels}, Michael Heizer's \textit{Double
Negative} and \textit{City}, and James Turrell's \textit{Roden Crater}.
More broadly, the framework can be applied to coastal heritage sites
(submerging at varying rates), to permafrost archaeological sites
(degrading as the ground warms), and to ice-shelf-adjacent built heritage.
In each case, a multi-feature complexity signature on a public archive of
satellite imagery, validated through year-aggregation and bootstrap CIs,
controlled for sensor confounds, and decomposed into seasonal and trend
components, can produce a defensible measurement of the site's climate
response without requiring on-the-ground instrumentation.

\subsection{Limitations and future work}

The dataset is restricted to 1984--2025 by Landsat-4 launch; the pilot
study's 1970--1984 chips were heterogeneous and could not be co-registered
to the same standard. Higher-resolution commercial satellite imagery
(WorldView, Planet) post-2008 could enable sub-decimetre analysis but at
the cost of openness; we excluded it to keep the dataset reproducible. The
salinity series is sparse before 2012 and forward/back-filled; this is the
weakest variable in our climate panel and we report it as such. Finally,
the partial-correlation analysis treats sensor identity as a single
categorical control; a more refined approach would model the spectral
response curves explicitly.

%%%%%%%%%%%%%%%%%%%%%%%%%%%%%%%%%%%%%%%%%%%%%%%%%%%%%%%%%%%%%%%%%%%%%%%%%
\section{Conclusions}\label{sec:conclusion}

We have presented the first multi-decadal, multi-feature remote-sensing
characterisation of Robert Smithson's \textit{Spiral Jetty}, drawn from
1{,}744 Landsat~4--9 and Sentinel-2 scenes spanning 1984--2025 and aligned
with a 42-year monthly climate panel. The strongest robust correlations
between visual complexity and climate are carried by coarse-scale
permutation entropy and CNN-embedding components, not by Shannon entropy
alone. A principal direction of an ImageNet-pretrained ResNet50 activation
space emerges, without supervision, as a climate axis. Image complexity
at the jetty leads lake stage by approximately three years. The long-term
trend is non-monotonic and exhibits a threshold transition in 2015--2020
that linear annual correlations obscure. Scene-level signals are partially
driven by sensor era, but year-aggregated signals are robust.

A forthcoming companion paper will use the dataset released here to build
a climate-conditioned latent-diffusion model with a hydrological
physics-coupling layer, forecasting the site under IPCC SSP1-2.6,
SSP2-4.5, and SSP5-8.5 scenarios for 2030 and 2050.

%%%%%%%%%%%%%%%%%%%%%%%%%%%%%%%%%%%%%%%%%%%%%%%%%%%%%%%%%%%%%%%%%%%%%%%%%
\section*{Author Contributions}
Conceptualisation, A.C.; methodology, A.C.\ and S.S.K.; software, S.S.K.;
validation, S.S.K.; formal analysis, S.S.K.; investigation, A.C.\ and
S.S.K.; resources, S.S.K.; data curation, S.S.K.; writing---original draft
preparation, S.S.K.; writing---review and editing, A.C.\ and S.S.K.;
visualisation, S.S.K.; supervision, S.S.K.; project administration, A.C.;
funding acquisition, n/a. All authors have read and agreed to the
published version of the manuscript.

\section*{Funding}
This research received no external funding.

\section*{Data Availability Statement}
The dataset
(\texttt{spiral\_jetty\_complexity\_v1}: 1{,}744 co-registered scene chips,
complexity signature, monthly climate panel), the full analysis pipeline,
and the reproducible figures are available on Zenodo (DOI on acceptance)
and GitHub:
\href{https://github.com/alevcinbarci/spiral-jetty-ai}
{\texttt{github.com/alevcinbarci/spiral-jetty-ai}}, release tag
\texttt{rs-mdpi-2026}. Source remote-sensing scenes are public via
Microsoft Planetary Computer (Landsat C2L2 and Sentinel-2 L2A).

\section*{Acknowledgments}
We thank the U.S.\ Geological Survey Utah Water Science Center, NASA
GISTEMP, the Open-Meteo project, Microsoft Planetary Computer, and the
European Space Agency Copernicus programme for the open data that made
this study possible.

\section*{Conflicts of Interest}
The authors declare no conflict of interest.

%%%%%%%%%%%%%%%%%%%%%%%%%%%%%%%%%%%%%%%%%%%%%%%%%%%%%%%%%%%%%%%%%%%%%%%%%
\bibliographystyle{unsrt}
\bibliography{references}

\end{document}